\documentclass[11pt]{article}

\usepackage[margin=1in]{geometry}
\usepackage[T1]{fontenc}
\usepackage{graphicx}
\usepackage{amsmath,amssymb,amsfonts}
\usepackage{booktabs}
\usepackage{enumerate}
\usepackage{pifont}
\usepackage{textcomp}
\usepackage{url}
\usepackage[hidelinks]{hyperref}

\providecommand{\grms}[2]{\begin{tabular}{@{}c@{}}$#1$\\$\pm#2$\end{tabular}}
\providecommand{\grmsb}[2]{\begin{tabular}{@{}c@{}}$\mathbf{#1}$\\$\mathbf{\pm#2}$\end{tabular}}
\providecommand{\grpm}[2]{$#1{\pm}#2$}
\providecommand{\grpmb}[2]{$\mathbf{#1{\pm}#2}$}
\providecommand{\grname}[2]{\begin{tabular}{@{}l@{}}#1\\#2\end{tabular}}

\newcommand{\cmark}{\ding{51}}
\newcommand{\xmark}{\ding{55}}

\begin{document}

\title{GRAC: Graph-Based Retrieval-Augmented Captioning for Surgical Scene Description}

\author{
Rochak Dhakal$^{1}$ \quad Chen Zhao$^{2}$ \quad Soup Cobb$^{3}$ \\[2pt]
Xin Qian$^{4}$ \quad Xiaofeng Yang$^{5}$ \quad Weihua Zhou$^{1,*}$ \\[6pt]
\small $^{1}$Department of Data Science, Michigan Technological University, Houghton, MI, USA \\
\small $^{2}$Department of Computer Science, Kennesaw State University, Marietta, GA, 30060, USA \\
\small $^{3}$Department of Computer Science, Michigan Technological University, Houghton, MI, USA \\
\small $^{4}$Department of Physics and Anatomy, Stony Brook University Hospital, Stony Brook, NY, USA \\
\small $^{5}$Department of Radiation and Cellular Oncology, University of Chicago, Chicago, IL, USA \\[6pt]
% \small \texttt{rochad@mtu.edu}, \texttt{czhao4@kennesaw.edu}, \texttt{bhcobb@mtu.edu}, \\
% \small \texttt{xin.qian@stonybrookmedicine.edu}, \texttt{xfyang@uchicago.edu}, \texttt{whzhou@mtu.edu} \\[2pt]
% \small $^{*}$Corresponding author: Weihua Zhou (\texttt{whzhou@mtu.edu})
}

\date{}

\maketitle

\begin{center}
\small Corresponding author: Weihua Zhou
(\texttt{whzhou@mtu.edu})
\end{center}

\begin{abstract}
\textbf{Purpose:} Surgical operative reports document the surgical procedures and support surgical training, performance assessment, and workflow review. However, extracting meaningful procedural information often requires extensive manual review and clinical expertise. Automated generation of clinically meaningful and contextually accurate surgical scene description can reduce manual review effort. Since many patterns recur across surgical scenes deep learning models have opportunity to learn these patterns integrating prior knowledge from scenes to generate more accurate and contextually relevant descriptions.

\textbf{Methods:} We propose GRAC (Graph-Based Retrieval-Augmented Captioning), a framework that represents surgical scenes as graphs to capture relationships among instruments, their actions, and anatomical structures. Given an input surgical scene graph, GRAC learns to retrieve relevant graph–caption pairs from the training data. It then compares the instrument–tissue interactions in the retrieved examples with those of the input scene and uses the resulting comparison features to condition retrieved-caption representations during generation.

\textbf{Results:} On the combined EndoVis-18 evaluation benchmark, GRAC achieved a CIDEr-D of 5.88, compared with 4.23 for SGT and 3.36 for the strongest conventional Transformer-based captioning baseline. GRAC also achieved an exact instrument--action--target triplet-set accuracy of 52.6\%, compared with 32.7\% for SGT and 26.2\% for the strongest conventional baseline. These findings demonstrate improvements in both linguistic quality and factual agreement. On the TORS dataset, GRAC achieved a CIDEr-D of 5.06, compared with 3.80 for the strongest baseline. 

\textbf{Conclusion:} GRAC demonstrates that integrating structured surgical scene representations with relevant information retrieved from prior examples enhances surgical scene understanding and generates more contextually meaningful descriptions.

\end{abstract}

\section{Introduction}\label{sec1}

Surgical video recordings provide valuable information about operative procedures and support surgical training, performance assessment, and workflow analysis~\cite{b1}. Automatically translating surgical activities into meaningful descriptions can facilitate procedural documentation, video retrieval, and the development of computer-assisted surgical systems. Surgical scene graphs provide a structured representation of these activities~\cite{b2}, where nodes represent surgical instruments and anatomical structures, and edges describe their relationships and interactions. 

Existing surgical report generation methods primarily learn a direct mapping from scene representations to natural-language descriptions~\cite{b3,b4,b5}. Although these approaches can generate descriptions of surgical activities, learning the relationships between structured scene information and language can be challenging when annotated training data are limited. Surgical procedures, however, contain recurring patterns of instrument use, actions, and anatomical interactions, often described using similar terminology. These characteristics suggest that descriptions of previously observed surgical scenes can provide useful prior information for generating new reports. Retrieval-based approaches offer a complementary direction~\cite{b6,b7} by incorporating relevant examples from the training data into the generation process.

\begin{figure}[!t]
\centering
\includegraphics[width=\linewidth]{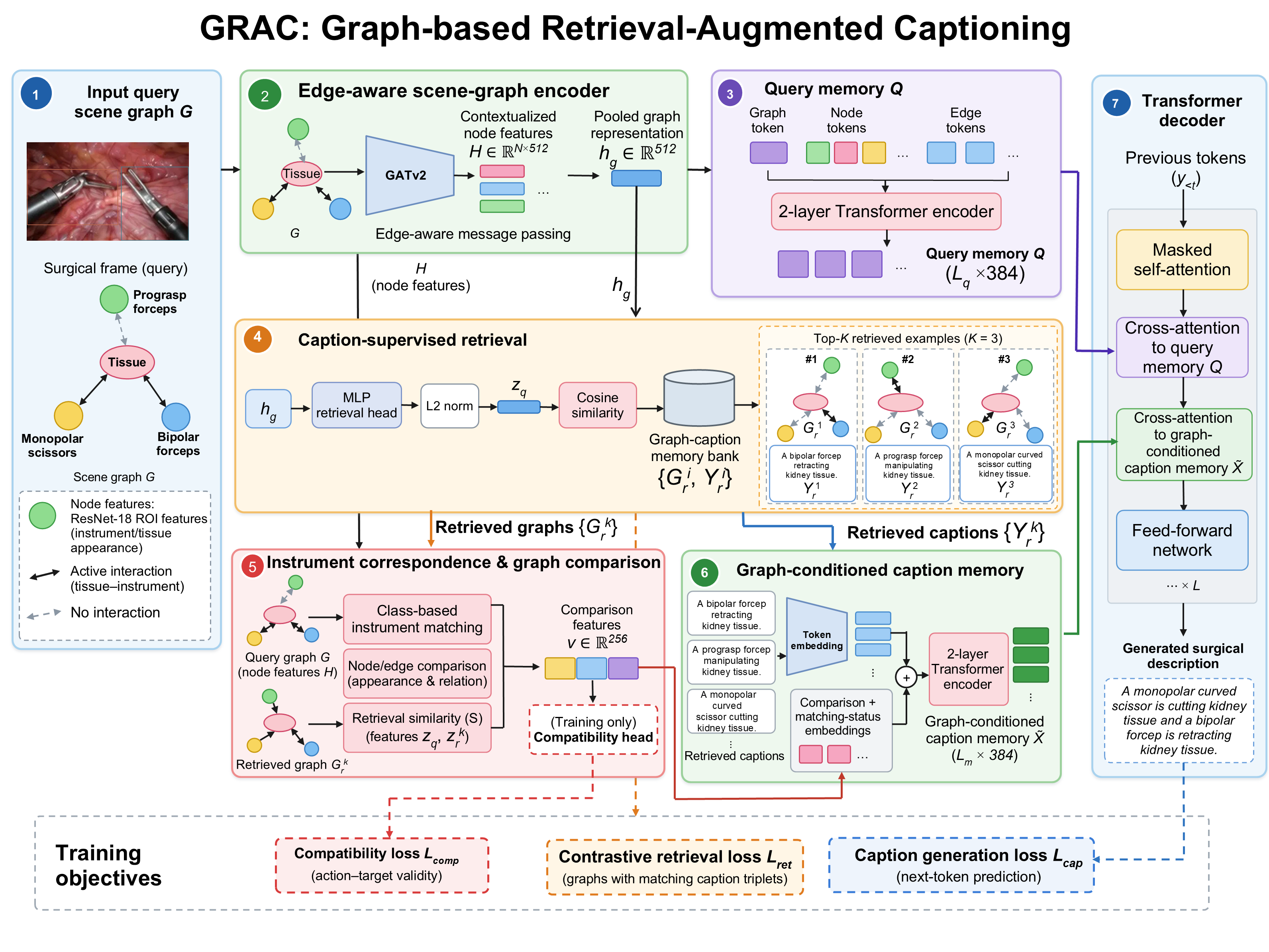}
\caption{Overview of GRAC. The input surgical scene graph is encoded to capture surgical interactions and retrieve similar prior scene graphs with their corresponding captions. GRAC then compares the retrieved graphs with the current scene and uses these comparisons to guide caption generation through a Transformer decoder.}
\label{fig1}
\end{figure}

Although surgical scenes frequently exhibit similar patterns, effectively using prior examples for caption generation requires careful consideration. Similar instruments may perform different actions or interact with different anatomical structures. Thus, retrieving relevant examples must be complemented by comparing their interactions with the current scene to ensure that the generated descriptions reflect the observed surgical activity. To address these challenges, we propose GRAC, a scene-graph-to-caption framework with three complementary components. First, an interaction-aware graph encoder represents relationships among surgical instruments, their actions, and anatomical structures. Second, a caption-supervised retrieval module learns to retrieve prior graph–caption pairs based on surgical activity rather than appearance alone. Third, a graph-conditioned caption memory compares the retrieved and current scenes and uses this comparison to guide caption generation. We evaluate GRAC on the EndoVis-18 and TORS datasets using language-generation metrics and measures of factual correctness.

The main contributions of this work are summarized as follows:

\begin{enumerate}[I.]
\item \textbf{Surgical activity-guided retrieval}  We introduce a caption-supervised contrastive learning strategy that brings graph embeddings of similar surgical activities closer together, enabling the model to retrieve relevant prior scene graphs and their corresponding captions based on learned graph similarity.
\item \textbf{Graph comparison-guided caption generation}: We develop an instrument-aligned comparison mechanism that captures differences between current and retrieved scene graphs and incorporates this information into retrieved caption representations to guide generation.
\item \textbf{Integrated surgical scene-captioning framework}: We combine interaction-aware graph encoding, activity-guided retrieval, and graph-conditioned caption generation in a unified framework, demonstrating their combined effectiveness through evaluations on EndoVis-18 and TORS and systematic ablation studies.
\end{enumerate}

\section{Related Work}\label{sec2}

\subsection{Surgical Captioning and Scene-Graph Methods}\label{subsec2.1}

Surgical scene understanding is central to context-aware computer-assisted intervention, where recognizing surgical instruments, anatomical structures, and their interactions provides important information for downstream applications. The EndoVis-18 robotic scene segmentation challenge~\cite{b8} has become a widely used benchmark for this setting. Building on this work, Islam et al.~\cite{b2} introduced surgical scene graphs, representing instruments and anatomical structures as nodes and their spatial or functional interactions as edges. These structured representations have subsequently been explored for surgical visual question answering~\cite{b9} and surgical report generation~\cite{b5,b10}.

Existing surgical captioning approaches have primarily focused on improving visual representations, graph encoding, and language generation. CIDA~\cite{b4} addresses domain shift and class-incremental adaptation, while SwinMLP-TranCAP~\cite{b11} reduces dependence on detector-based regional features through shifted-window MLP patches. Scene Graph-guided Transformer (SGT)~\cite{b5} incorporates graph relationships into Transformer-based generation to capture instrument--tissue interactions. Its extension, SGT++~\cite{b10}, further develops graph-based representations by modeling surgical relationships through explicit and implicit interactions. These approaches demonstrate the value of structured surgical information for generating descriptions of operative activities.

However, existing methods primarily learn surgical language patterns through model parameters, despite the repetitive and structured nature of surgical descriptions. Previously observed captions may provide useful examples of how similar activities are described, particularly when annotated training data are limited. GRAC extends scene-graph-based captioning by incorporating retrieved graph--caption examples as semantic priors, allowing the generator to draw on previously observed descriptions in addition to the current scene representation.

Transformer and Transformer-based image-captioning architectures~\cite{b12,b13,b14,b15} provide complementary approaches to sequence generation. For graph encoding, GRAC employs GATv2~\cite{b16}, which introduces dynamic attention to address limitations of the static attention ranking in GAT~\cite{b17}. This enables context-dependent aggregation of graph information, which is relevant when surgical scenes contain similar entities but different interaction patterns.

\subsection{Retrieval-Augmented Generation for Captioning}\label{subsec2.2}

Retrieval-augmented generation (RAG)~\cite{b18} combines a parametric generator with an external memory, enabling previously observed information to guide the generation of new content. In image captioning, approaches such as SmallCap~\cite{b6}, EXTRA~\cite{b7}, and COS-Net~\cite{b19} explore the use of retrieved captions or semantic information to support description generation. Retrieval has also attracted interest in medical reporting, where descriptions often follow recurring linguistic patterns while requiring accurate representation of clinical findings. X-REM~\cite{b20} uses contrastive image--text matching to retrieve relevant chest X-ray reports, while FactMM-RAG~\cite{b21} incorporates fact-aware retrieval to support factual consistency in radiology report generation.

Liu et al.~\cite{b22} explored retrieval-augmented surgical captioning using a large multimodal model that retrieves surgical concepts from visually similar images to support long-form description generation. Their approach incorporates retrieved knowledge into visual representations through a dedicated attention mechanism. In contrast, GRAC operates on structured surgical scene graphs and learns to retrieve prior graph--caption pairs through caption-supervised training. It further compares instrument-level interactions between the retrieved and current scenes to condition caption generation on their relational similarities and differences.

These studies demonstrate the potential of retrieval-based approaches to incorporate prior information into surgical caption generation. However, surgical activities are defined by specific relationships among instruments, actions, and anatomical targets, which may differ even between otherwise similar scenes. Consequently, the relevance of a retrieved description depends not only on overall scene similarity but also on whether its underlying interactions correspond to those of the current scene. GRAC addresses this distinction at two levels: caption-supervised retrieval selects examples according to surgical activity, and graph-conditioned caption memory grounds the retrieved text in the interactions of the current scene. Through this integration, retrieved descriptions provide contextual information that is informed by graph-based relational agreement with the current scene, connecting structured scene understanding with retrieval-guided caption generation.

\section{Methodology}\label{sec3}

\subsection{Problem Formulation and Overview}\label{subsec3.1}

GRAC generates a natural-language description of surgical activity from a scene graph representing a single video frame. Following the surgical scene-graph formulation of Islam et al.~\cite{b2}, each frame is represented as $G = (V,E,X^{v},X^{e})$, where $V$ contains instrument nodes and a single tissue node, and $E$ represents their connections. The feature matrices $X^{v} \in \mathbb{R}^{|V| \times 512}$ and $X^{e} \in \mathbb{R}^{|E| \times 512}$ contain ResNet-18 ROI and interaction features extracted from annotated bounding boxes using the representation of Xu et al.~\cite{b3}. Nonzero edge features represent annotated active (non-idle) tissue--instrument interactions without explicitly specifying the performed action, while all other connections have zero-valued edge features. 

As illustrated in Fig.~\ref{fig1}, given a query graph $G$, the objective is to generate a caption $Y = (y_{1},\ldots,y_{T})$ describing the observed surgical activity. The caption's factual content is represented by a set $\mathcal{S}(Y)$ of instrument--action--target triplets $\langle i,a,t \rangle$. GRAC maintains a training-only memory bank $\mathcal{B} = \{ (G_{j},Y_{j}) \}_{j = 1}^{N'}$, where each entry associates a training scene graph with its reference caption. For a query graph ($G$), GRAC selects $K = 3$ graph--caption examples, denoted by $\mathcal{R}(G) = \{ (G_{r_{k}},Y_{r_{k}}) \}_{k = 1}^{K} \subseteq \mathcal{B}$, where ($r_{k}$) indexes the ($k$\textsuperscript{th}) retrieved example. Each selected graph--caption pair $(G_{r_{k}},Y_{r_{k}})$ is referred to as a retrieved example, consisting of a retrieved graph and its associated caption. During training, retrieval candidates from the query's own surgical sequence are excluded to prevent self-retrieval and near-duplicate matches. Then, the caption generation is formulated as $P_{\theta}(Y \mid G,\mathcal{R}(G))$. During inference, the full training memory bank is available for retrieval, as the query originates from an unseen surgical sequence.

\subsection{Scene-graph encoder and query memory}\label{subsec3.2}

Given a scene graph $G = (V,E)$ with $N = |V|$ nodes representing the tissue and visible surgical instruments, GRAC employs an edge-aware Graph Attention Network (GATv2)~\cite{b16} to capture structural and relational information.  The encoder consists of two GATv2 layers, each with four attention heads of dimension 128, producing 512-dimensional node representations. During message passing, node features are updated through attention-weighted aggregation of neighboring node features, with edge information guiding the aggregation. Edge features contribute to the computation of attention coefficients, allowing node representations to be updated according to the observed instrument--tissue interactions. The resulting node features $H \in \mathbb{R}^{N \times 512}$ are contextualized with edge information and are aggregated through attention pooling to obtain a graph-level representation $h_{g} \in \mathbb{R}^{512}$.

To preserve global scene context and individual instrument–tissue interactions, GRAC constructs a query memory using three types of tokens: a graph-level token ($h_g$), contextualized node tokens ($H$), and interaction tokens derived from the original 512-dimensional edge features. Each node contributes one token, while each annotated instrument–tissue interaction contributes two directed edge tokens. The sequence is padded to a fixed length of 19 tokens (one graph token, up to six node tokens, and twelve edge tokens), sufficient to accommodate all graphs in both EndoVis-18 and TORS without truncation, with padded positions masked during attention. All tokens are projected into a shared 384-dimensional space and augmented with learned token-type and positional embeddings. A learned $[\mathrm{CLS}]$ token is prepended and the sequence is processed by a two-layer Transformer encoder with four attention heads to produce the query memory ($Q$). This memory, together with the graph-conditioned representations of retrieved captions, guides the report decoder in generating surgical descriptions.

\subsection{Contrastive Caption Retrieval}\label{subsec3.3}

GRAC retrieves reference examples whose captions describe surgical activities similar to the query scene. During training, caption-supervised contrastive learning establishes a shared embedding space in which training graphs associated with similar relationships are encouraged to have similar representations. A query graph is subsequently mapped into this learned space using the same graph encoder and retrieval head, allowing GRAC to identify relevant training graphs through embedding similarity and retrieve their associated captions.

Here, a retrieval head $f_{ret}$ is implemented as a two-layer multilayer perceptron (MLP) with GELU activation and maps the pooled graph representation $h_{g}$ of the current frame to an L2-normalized embedding
\begin{equation}
\mathbf{z} = \frac{f_{ret}(\mathbf{h}_{g})}{\left\| f_{ret}(\mathbf{h}_{g}) \right\|_{2}} \in \mathbb{R}^{128}
\label{eq1}
\end{equation}

For each query graph, its L2-normalized embedding is compared with the stored training graph embeddings using cosine similarity. During training, candidates from the query's own surgical sequence are excluded to prevent self-retrieval and near-duplicate matches. Candidates are ranked by similarity, and the highest-ranked scene from each distinct instrument–action–target triplet set is retained until three graph–caption pairs are selected. This promotes diversity while reducing redundant descriptions of similar activities. Although InfoNCE learns graph similarities, the top-$K$ selection is non-differentiable, and similarity scores are detached before graph comparison. Thus, the retrieval head is trained only through InfoNCE, while the shared graph encoder is jointly optimized through caption-generation, contrastive retrieval, and compatibility objectives. The retrieved captions provide contextual examples, then GRAC utilizes subsequent graph comparison to account for differences between the query and retrieved surgical interactions.

\subsection{Graph Comparison with Retrieved Scenes}\label{subsec3.4}

For each retrieved scene, GRAC identifies instruments corresponding to those in the query graph and compares their instrument–tissue interactions. Instrument correspondence is established based on class labels, obtained from training annotations for retrieved graphs and predicted using a frozen eight-class multinomial logistic regression classifier for query graphs. For each matched instrument pair, GRAC extracts their contextualized node representations $h_q$ and $h_r$ from the GATv2 node embeddings $H$ of the query and retrieved graphs, respectively, along with their corresponding raw instrument–tissue edge features $e_q$ and $e_r$. Two learned linear projections $P,R:\mathbb{R}^{512}\rightarrow\mathbb{R}^{128}$ transform the node and edge features, respectively. With $q=P(h_q)$ and $r=P(h_r)$, the comparison vector is computed as

\begin{equation}
v = f_{cmp}\left( \left[ q;r;q-r;q\odot r;R(e_q);R(e_r);R(e_q-e_r);s \right] \right) \in \mathbb{R}^{256}
\label{eq3}
\end{equation}

where $\odot$ denotes element-wise multiplication, $s$ is the detached cosine similarity score from retrieval, and $f_{cmp}$ consists of two linear layers (897$\rightarrow$256$\rightarrow$256), GELU activation, and layer normalization. The terms $q$ and $r$ preserve individual instrument representations, while $q-r$ captures their differences and $q\odot r$ captures feature-wise agreement. Similarly, $R(e_q)$ and $R(e_r)$ represent the respective instrument–tissue interactions, while $R(e_q-e_r)$ captures their differences. Together with the graph-level similarity score $s$, these features form a comparison vector that conditions the corresponding retrieved caption tokens, as described in Section~\ref{subsec3.5}.

\subsection{Graph-Conditioned Caption Representations}\label{subsec3.5}

For each query, GRAC constructs a caption memory from three retrieved examples, each padded to 20 word-token slots. Each caption is processed by a two-layer Transformer encoder (Table~\ref{tab6}) to obtain contextualized token representations. The comparison vectors obtained in Section~\ref{subsec3.4} are then incorporated into the corresponding caption tokens to reflect differences between the retrieved and current surgical scenes.

For each token $l$, the associated comparison vectors are averaged to obtain $\overline{\mathbf{v}}_l\in\mathbb{R}^{256}$, or set to zero if no matching instrument is available. A learned embedding $\mathbf{m}_l\in\mathbb{R}^{32}$ indicates its matching status (neutral, matched, shared, ambiguous, or unmatched). These features are combined with the original caption-token representation $\mathbf{x}_l$ as

\begin{equation}
\widetilde{\mathbf{x}}_{l} = LN\left( W_{h}\mathbf{x}_{l} + W_{u}\left[ \overline{\mathbf{v}}_{l};\mathbf{m}_{l} \right] \right)
\label{eq4}
\end{equation}

\noindent where $W_h$ projects the caption-token representation into the 384-dimensional decoder space, $W_u\in\mathbb{R}^{384\times288}$ projects the comparison and matching-status features, and $LN$ denotes layer normalization. The resulting graph-conditioned caption memory, together with the independently encoded query graph, guides surgical description generation.

\subsection{Training Objective and Inference}\label{subsec3.6}

GRAC is jointly trained with three objectives: caption generation for accurate surgical descriptions, contrastive retrieval learning to align scenes with similar surgical activities, and compatibility supervision to capture factual agreements and discrepancies between retrieved captions and the query scene.

\textbf{Caption-generation loss.} The decoder is trained using teacher forcing with token-level cross-entropy and label smoothing. Given a ground-truth caption $Y$, the caption-generation loss is
\begin{equation}
\mathcal{L}_{cap} = -\sum_{t}\sum_{w \in \mathcal{V}}\widetilde{p}_{t}(w)\log P_{\theta}\left( w \mid Y_{<t},Q,\widetilde{X} \right)
\label{eq5}
\end{equation}
where $\mathcal{V}$ is the caption vocabulary, $\widetilde{p}_{t}$ is the label-smoothed target distribution, $Q$ is the query memory, and $\widetilde{X}$ denotes the graph-conditioned retrieved caption representations.

\textbf{Contrastive retrieval loss.} GRAC employs a multi-positive InfoNCE objective~\cite{b23} to encourage similar graph embeddings for identical surgical activities across sequences while distinguishing different activities, even when the same instruments are present. Each batch contains 64 training frames: 16 forming eight cross-sequence positive pairs, 16 instrument-matched hard negatives, and 32 randomly sampled frames. For any two frames $i$ and $j$, the temperature-scaled similarity between their L2-normalized graph embeddings $\mathbf{z}_i$ and $\mathbf{z}_j$ is $s_{ij}=\mathbf{z}_i^\top\mathbf{z}_j/\tau$, where $\tau=0.1$. The contrastive loss is defined as

\begin{equation}
\mathcal{L}_{ret} =
-\frac{1}{|\mathcal{I}|}
\sum_{i\in\mathcal{I}}
\frac{1}{|\mathcal{P}(i)|}
\sum_{p\in\mathcal{P}(i)}
\log
\frac{\exp(s_{ip})}
{\sum_{a\in\mathcal{A}(i)}\exp(s_{ia})}
\label{eq2}
\end{equation}

where $\mathcal{I}$ is the set of frames $i$ having at least one positive match, $\mathcal{P}(i)$ contains positive frames $p$ with identical instrument--action--target triplet sets from different sequences, and $\mathcal{A}(i)$ contains comparison frames $a$, excluding $i$ and frames sharing both its triplet set and surgical sequence. The notation $|\cdot|$ denotes the number of elements in a set. Frames without positive matches do not contribute individual loss terms but can still serve as negatives for other frames.

\textbf{Compatibility supervision.} An auxiliary compatibility head learns whether the action–target relationship in a retrieved caption agrees with the corresponding instrument in the query scene. Using graph-comparison features and learned action–target embeddings, it predicts a compatibility score $\kappa_r$. Labels are derived from the query annotations, with additional negative examples generated from counterfactual action–target pairs. The objective uses positive-class-weighted binary cross-entropy:

\begin{equation}
\mathcal{L}_{comp} = -\frac{1}{\widetilde{N}}\sum_{r=1}^{\widetilde{N}}
\left[
w y_r\log\sigma(\kappa_r)
+(1-y_r)\log(1-\sigma(\kappa_r))
\right]
\label{eq7}
\end{equation}

\noindent where $y_r$ is the binary compatibility label, $\widetilde{N}$ is the number of scored pairs, and $w=\widetilde{N}_{neg}/\widetilde{N}_{pos}$ is the positive-class weight estimated from initial training selections. This auxiliary loss supervises graph-comparison representations during training, while the compatibility head is omitted during inference.
Thus, the overall training objective is
\begin{equation}
\mathcal{L} = \mathcal{L}_{cap}
+ \lambda_{ret}\mathcal{L}_{ret}
+ \lambda_{comp}\mathcal{L}_{comp}
\label{eq8}
\end{equation}

The three objectives jointly optimize caption generation, graph retrieval, and interaction-level compatibility. The captioning loss supervises the generated descriptions, the contrastive retrieval loss encourages the selection of relevant prior examples, and the compatibility loss supervises agreement between current and retrieved interactions. These objectives update their respective modules and shared representations during training. Since retrieval involves discrete selection of ranked examples, gradients from the captioning loss do not propagate through the retrieval ranking. The shared graph encoder, however, receives supervision from caption generation, contrastive retrieval, and compatibility objectives.

\textbf{Inference.} Given an input surgical scene graph, GRAC encodes its instrument--tissue interactions and retrieves ($K=3$) examples with distinct caption triplet sets from the fixed training memory bank. The retrieved graphs are compared with the query graph, and the resulting features condition their associated caption representations. The decoder then generates a description using both the encoded query graph and the graph-conditioned caption memory. The auxiliary compatibility head is used only during training. Captions are generated using beam search with a beam width of 2, a length-normalization factor of 0.8, and a maximum output length of 20 tokens.

\section{Results}\label{sec5}

\subsection{Comparison with Baselines}\label{subsec5.1}

\begin{figure}[htbp]
\centering
\includegraphics[width=\textwidth]{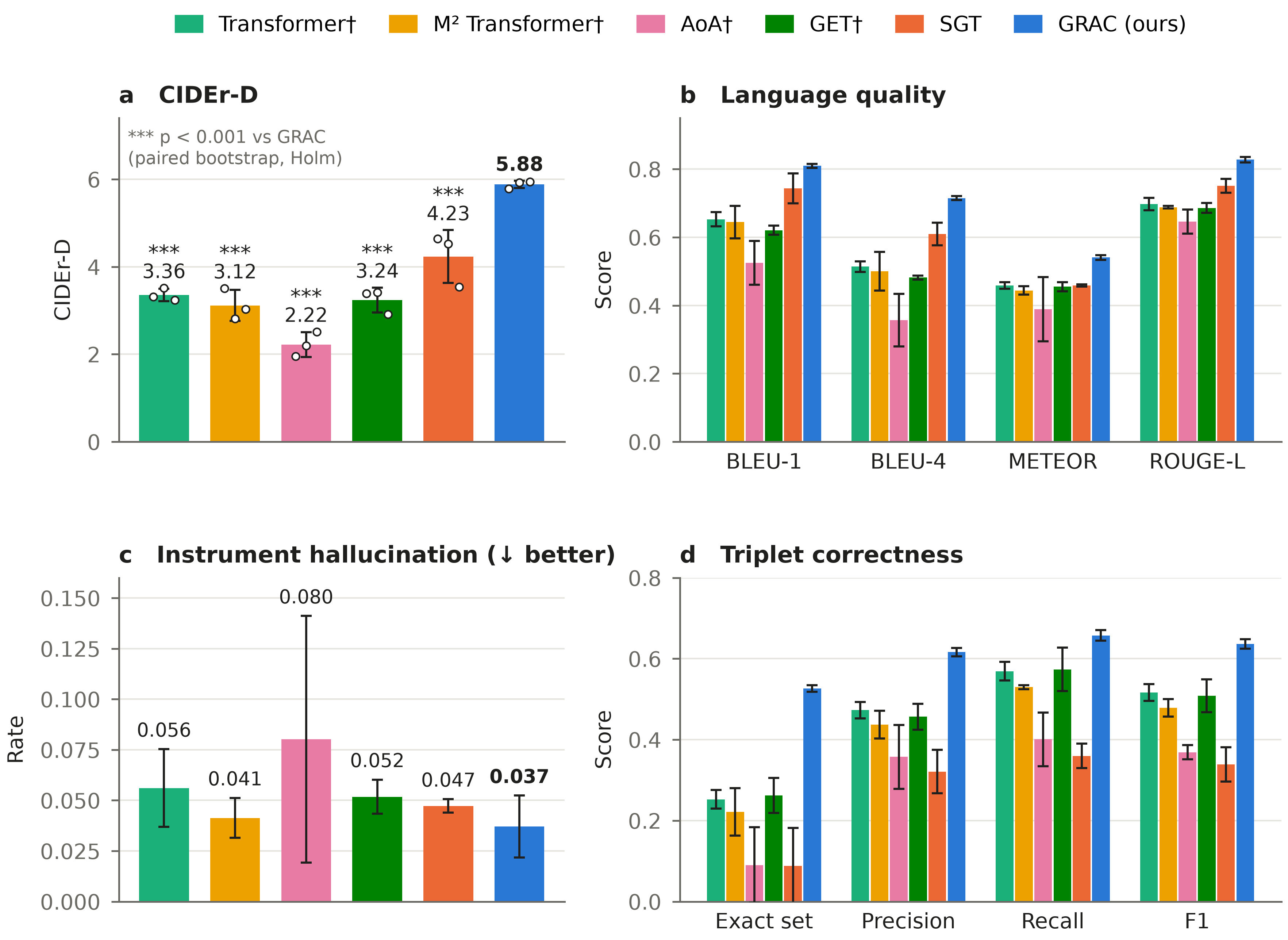}
\caption{Comparison of GRAC with interaction-feature-augmented baselines (\textdagger) and SGT on EndoVis-18 (392 frames). Bars show the mean over three seeds, with error bars for SD; in (a), circles are individual seeds. (a) CIDEr-D; *** p $<$ 0.001 vs GRAC (paired frame-level bootstrap, B = 10,000, Holm-corrected). (b) Language metrics. (c) Instrument hallucination (lower is better). (d) Triplet-level factual correctness.}\label{fig2}
\end{figure}

GRAC achieves the highest reported language-generation performance among the evaluated methods on EndoVis-18, with CIDEr-D of 5.883 $\pm$ 0.085, BLEU-4 of 0.715 $\pm$ 0.006, METEOR of 0.541 $\pm$ 0.007, and ROUGE-L of 0.828 $\pm$ 0.008 (Table~\ref{tab1}, Fig.~\ref{fig2}). SGT achieves the second-highest CIDEr-D of 4.234 $\pm$ 0.604, while Transformer is the strongest conventional baseline on this measure at 3.356 $\pm$ 0.144.

For triplet-based factual agreement (Table~\ref{tab2}), GRAC achieves an exact triplet-set accuracy of 0.526 $\pm$ 0.008 and a triplet F1 of 0.636 $\pm$ 0.012. SGT is the strongest baseline in exact-set accuracy (0.327 $\pm$ 0.114), followed by GET (0.262 $\pm$ 0.043). SGT achieves a triplet F1 of 0.510 $\pm$ 0.107. GRAC exceeds SGT by 0.199 in exact-set accuracy (95\% CI [0.171, 0.228]) and 0.126 in triplet F1 (95\% CI [0.107, 0.147]).

Paired frame-level bootstrap analysis with 10,000 resamples shows that GRAC outperforms all baselines in CIDEr-D, exact triplet-set accuracy, and triplet F1 (Holm-adjusted p $<$ 0.001). GRAC exceeds every baseline in each of the three seeds on these primary metrics. An error analysis further examined instruments, action, and target correctness. Each predicted triplet was classified as having a wrong or extra instrument relative to the reference, a wrong action, a wrong target, or no error. Wrong-instrument rates were calculated over all predicted triplets, whereas wrong-action and wrong-target rates were calculated among predictions whose instruments appeared in the reference. GRAC had fewer wrong or extra instruments (0.210) than SGT (0.260) and the conventional baselines (0.374--0.439). Among instrument-matched predictions, GRAC also had a lower wrong-action rate (0.215) than SGT (0.344), AoA (0.343), and Transformer (0.242), with statistically significant differences (Holm-adjusted p $\leq$ 0.034). Differences from M\textsuperscript{2} Transformer and GET were not significant. Wrong-target rates remained below 0.02 across all models under the normalized parser.

\begin{table}[!htbp]
\caption{Performance on EndoVis-18 (392 frames). (a) Language-generation metrics; values are mean (first line) and $\pm$SD (second line) over three seeds, and higher is better for all metrics. (b) Paired frame-level bootstrap comparison of GRAC with each baseline in CIDEr-D (GRAC minus baseline; 10,000 resamples; two-sided, Holm-adjusted p-values).}\label{tab1}
\centering
\footnotesize
\setlength{\tabcolsep}{2.6pt}
\renewcommand{\arraystretch}{1.15}
\begin{tabular*}{\textwidth}{@{\extracolsep{\fill}}l*{7}{c}@{}}
\toprule
\multicolumn{8}{@{}l}{\textit{(a) Language-generation metrics}} \\
\midrule
& \multicolumn{4}{c}{\textbf{BLEU}} & & & \\
\cmidrule(lr){2-5}
\textbf{Model} & \textbf{1} & \textbf{2} & \textbf{3} & \textbf{4} & \textbf{METEOR} & \textbf{ROUGE-L} & \textbf{CIDEr-D} \\
\midrule
Transformer$^\dagger$ & \grms{0.653}{0.021} & \grms{0.604}{0.019} & \grms{0.556}{0.017} & \grms{0.514}{0.016} & \grms{0.458}{0.009} & \grms{0.697}{0.018} & \grms{3.356}{0.144} \\
\addlinespace[2pt]
\grname{M$^{2}$}{Transformer$^\dagger$} & \grms{0.645}{0.048} & \grms{0.591}{0.047} & \grms{0.541}{0.051} & \grms{0.501}{0.057} & \grms{0.444}{0.012} & \grms{0.688}{0.004} & \grms{3.116}{0.356} \\
\addlinespace[2pt]
AoA$^\dagger$ & \grms{0.525}{0.064} & \grms{0.466}{0.069} & \grms{0.409}{0.070} & \grms{0.357}{0.077} & \grms{0.389}{0.094} & \grms{0.646}{0.035} & \grms{2.220}{0.283} \\
\addlinespace[2pt]
GET$^\dagger$ & \grms{0.621}{0.013} & \grms{0.570}{0.010} & \grms{0.523}{0.007} & \grms{0.482}{0.006} & \grms{0.455}{0.014} & \grms{0.686}{0.014} & \grms{3.239}{0.286} \\
\addlinespace[2pt]
SGT & \grms{0.744}{0.044} & \grms{0.694}{0.039} & \grms{0.647}{0.034} & \grms{0.610}{0.033} & \grms{0.459}{0.003} & \grms{0.751}{0.021} & \grms{4.234}{0.604} \\
\addlinespace[2pt]
\textbf{GRAC (ours)} & \grmsb{0.810}{0.006} & \grmsb{0.774}{0.006} & \grmsb{0.741}{0.006} & \grmsb{0.715}{0.006} & \grmsb{0.541}{0.007} & \grmsb{0.828}{0.008} & \grmsb{5.883}{0.085} \\
\bottomrule
\end{tabular*}
\par\vspace{8pt}
\begin{tabular}{@{}lccc@{}}
\toprule
\multicolumn{4}{@{}l}{\textit{(b) CIDEr-D comparison with GRAC}} \\
\midrule
\textbf{Baseline} & \textbf{$\Delta$CIDEr-D} & \textbf{95\% CI} & \begin{tabular}[b]{@{}c@{}}\textbf{Adjusted}\\\textbf{p-value}\end{tabular} \\
\midrule
Transformer$^\dagger$ & $+2.527$ & $[2.134,\,2.932]$ & $<0.001$ \\
M$^{2}$ Transformer$^\dagger$ & $+2.768$ & $[2.338,\,3.216]$ & $<0.001$ \\
AoA$^\dagger$ & $+3.663$ & $[3.298,\,4.044]$ & $<0.001$ \\
GET$^\dagger$ & $+2.644$ & $[2.252,\,3.048]$ & $<0.001$ \\
SGT & $+1.650$ & $[1.394,\,1.917]$ & $<0.001$ \\
\bottomrule
\end{tabular}
\par\vspace{3pt}\raggedright\scriptsize
$^\dagger$ Baseline augmented with GRAC's interaction-edge features: the original models receive only node features, and we added one input token per instrument–tissue interaction without other architectural changes. 
\end{table}

\begin{table}[!htbp]
\caption{Triplet-based factual agreement on EndoVis-18 (392 frames). Values are mean $\pm$ SD over three seeds. Higher is better except instrument hallucination ($\downarrow$).}\label{tab2}
\centering
\footnotesize
\setlength{\tabcolsep}{2.6pt}
\renewcommand{\arraystretch}{1.15}
\begin{tabular*}{\textwidth}{@{\extracolsep{\fill}}lccccc@{}}
\toprule
\textbf{Model} & \begin{tabular}[b]{@{}c@{}}\textbf{Exact}\\\textbf{set}\end{tabular} & \begin{tabular}[b]{@{}c@{}}\textbf{Triplet}\\\textbf{prec.}\end{tabular} & \begin{tabular}[b]{@{}c@{}}\textbf{Triplet}\\\textbf{recall}\end{tabular} & \begin{tabular}[b]{@{}c@{}}\textbf{Triplet}\\\textbf{F1}\end{tabular} & \begin{tabular}[b]{@{}c@{}}\textbf{Instr.}\\\textbf{halluc. $\downarrow$}\end{tabular} \\
\midrule
Transformer$^\dagger$ & \grpm{0.253}{0.023} & \grpm{0.473}{0.020} & \grpm{0.569}{0.023} & \grpm{0.516}{0.021} & \grpm{0.056}{0.019} \\
M$^{2}$ Transformer$^\dagger$ & \grpm{0.221}{0.059} & \grpm{0.437}{0.034} & \grpm{0.530}{0.005} & \grpm{0.479}{0.022} & \grpm{0.041}{0.010} \\
AoA$^\dagger$ & \grpm{0.090}{0.093} & \grpm{0.357}{0.079} & \grpm{0.401}{0.066} & \grpm{0.369}{0.017} & \grpm{0.080}{0.061} \\
GET$^\dagger$ & \grpm{0.262}{0.043} & \grpm{0.457}{0.032} & \grpm{0.573}{0.054} & \grpm{0.508}{0.041} & \grpm{0.052}{0.008} \\
SGT & \grpm{0.327}{0.114} & \grpm{0.479}{0.083} & \grpm{0.547}{0.137} & \grpm{0.510}{0.107} & \grpm{0.047}{0.003} \\
\textbf{GRAC (ours)} & \grpmb{0.526}{0.008} & \grpmb{0.616}{0.010} & \grpmb{0.657}{0.013} & \grpmb{0.636}{0.012} & \grpmb{0.037}{0.015} \\
\bottomrule
\end{tabular*}
\end{table}

\subsection{Ablation Study}\label{subsec5.2}

\textbf{Retrieval-set size.} We evaluated the effect of retrieving K $\in$ \{1, 3, 5\} examples on the development split across three seeds. Although caption quality was comparable for K = 1 and K = 3, retrieving three examples increased the likelihood of including the correct surgical activity (R@1 = 0.621; R@3 = 0.855). Increasing the retrieval set to K = 5 did not improve caption quality. We therefore retained the prespecified $K=3$ configuration to provide broader coverage of potentially relevant surgical activities.

\textbf{Interaction Features and Retrieval}. The ablation results demonstrate the complementary contributions of interaction-edge features and caption retrieval. Removing both components yields a CIDEr-D score of 4.870. Incorporating interaction-edge features without retrieval improves performance to 5.632, highlighting the importance of explicitly modeling instrument--tissue relationships. Adding caption-only retrieval further increases CIDEr-D to 5.805, suggesting that retrieved descriptions provide complementary contextual information beyond the query graph representation. Together, these findings indicate that relational scene modeling and retrieval augmentation contribute incrementally to caption generation performance.

\textbf{Graph Comparison and Compatibility Supervision.} Introducing graph comparison without compatibility supervision yields a CIDEr-D score of 5.744, compared with 5.805 for caption-only retrieval, indicating that comparison features alone do not improve mean caption quality under the evaluated configuration. Incorporating compatibility supervision increases CIDEr-D to 5.883, exact triplet-set accuracy to 0.526, and triplet F1 to 0.636, compared with 5.805, 0.520, and 0.630, respectively, for caption-only retrieval. However, these incremental differences do not reach statistical significance in the paired frame-level analysis (CIDEr-D: p = 0.054; exact triplet-set accuracy: p = 0.21; triplet F1: p = 0.23). In contrast, instrument hallucination decreases from 0.049 to 0.037 (p = 0.003). These findings suggest that compatibility supervision may help the model utilize graph-comparison features, with the clearest observed benefit being reduced instrument hallucination rather than a substantial improvement in aggregate caption accuracy.

This result suggests that introducing relational comparison features does not necessarily improve generation when their relevance to the current surgical activity is learned only through caption supervision. The improvement observed with compatibility supervision is consistent with the auxiliary objective helping the model learn more useful comparison representations, although the present experiments do not directly establish this mechanism.

\textbf{Temporal Context.} Incorporating neighboring frames ($t-1$), ($t$), and ($t+1$) yields a CIDEr-D score of 5.749 and an exact triplet-set accuracy of 0.514, compared with 5.883 and 0.526, respectively, for the single-frame GRAC model. Although temporal context produces a slight improvement in BLEU-4, it does not improve CIDEr-D or exact triplet-set accuracy. These findings indicate that incorporating adjacent frames provides no additional benefit on the primary evaluation metrics under the current experimental setting. The single-frame configuration was selected on the DEV split and does not require neighboring frames during inference; benchmark results for the temporal variant are reported for completeness.

\begin{figure}[htbp]
\centering
\includegraphics[width=\textwidth]{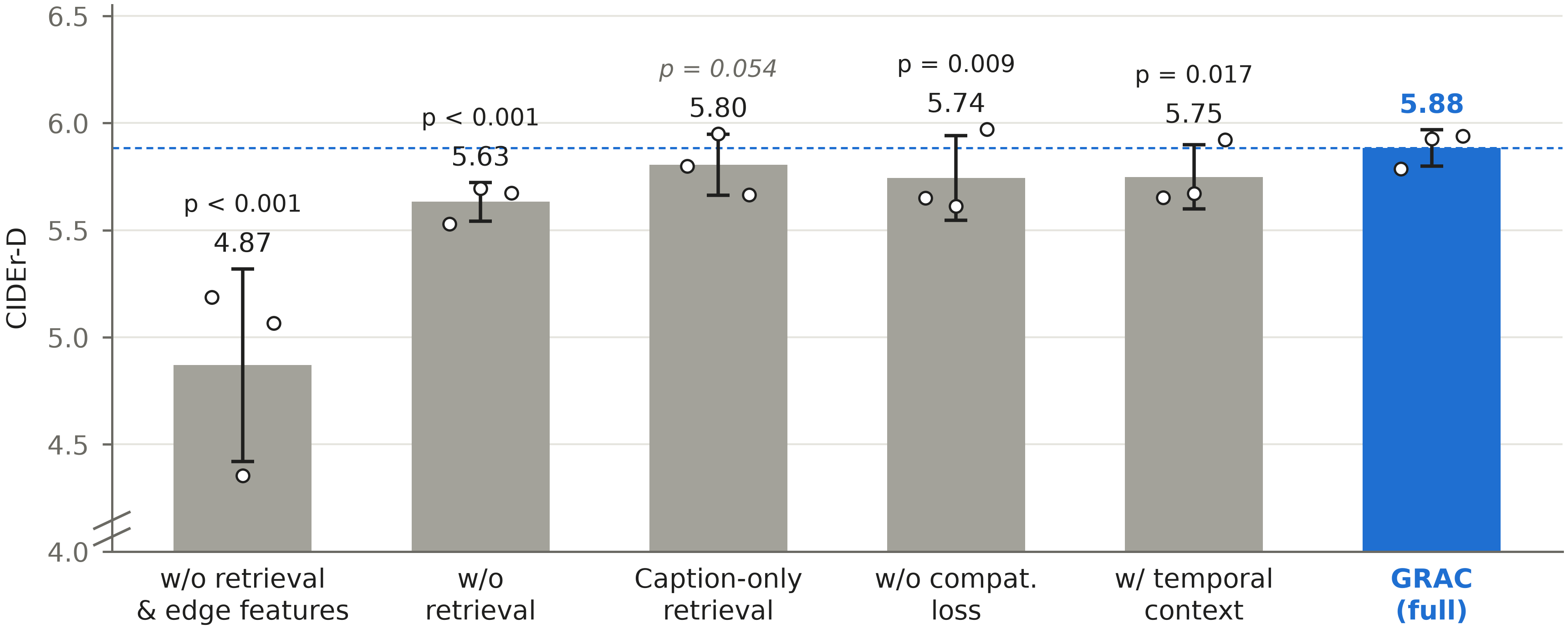}
\caption{Ablation of GRAC on EndoVis-18 (392 frames). Bars show CIDEr-D (mean $\pm$ SD over three seeds; dots are individual seeds; y-axis starts at 4.0). p-values compare each variant with GRAC using a paired frame-level bootstrap (B = 10,000), Holm-corrected across comparisons.}\label{fig3}
\end{figure}

Table~\ref{tab3} and Fig.~\ref{fig3} summarize the language-generation performance of the ablation variants across three training seeds. Paired frame-level bootstrap comparisons were performed between full GRAC and the ablation variants, whereas the successive component differences were otherwise interpreted descriptively.

\begin{figure}[htbp]
\centering
\includegraphics[width=\textwidth]{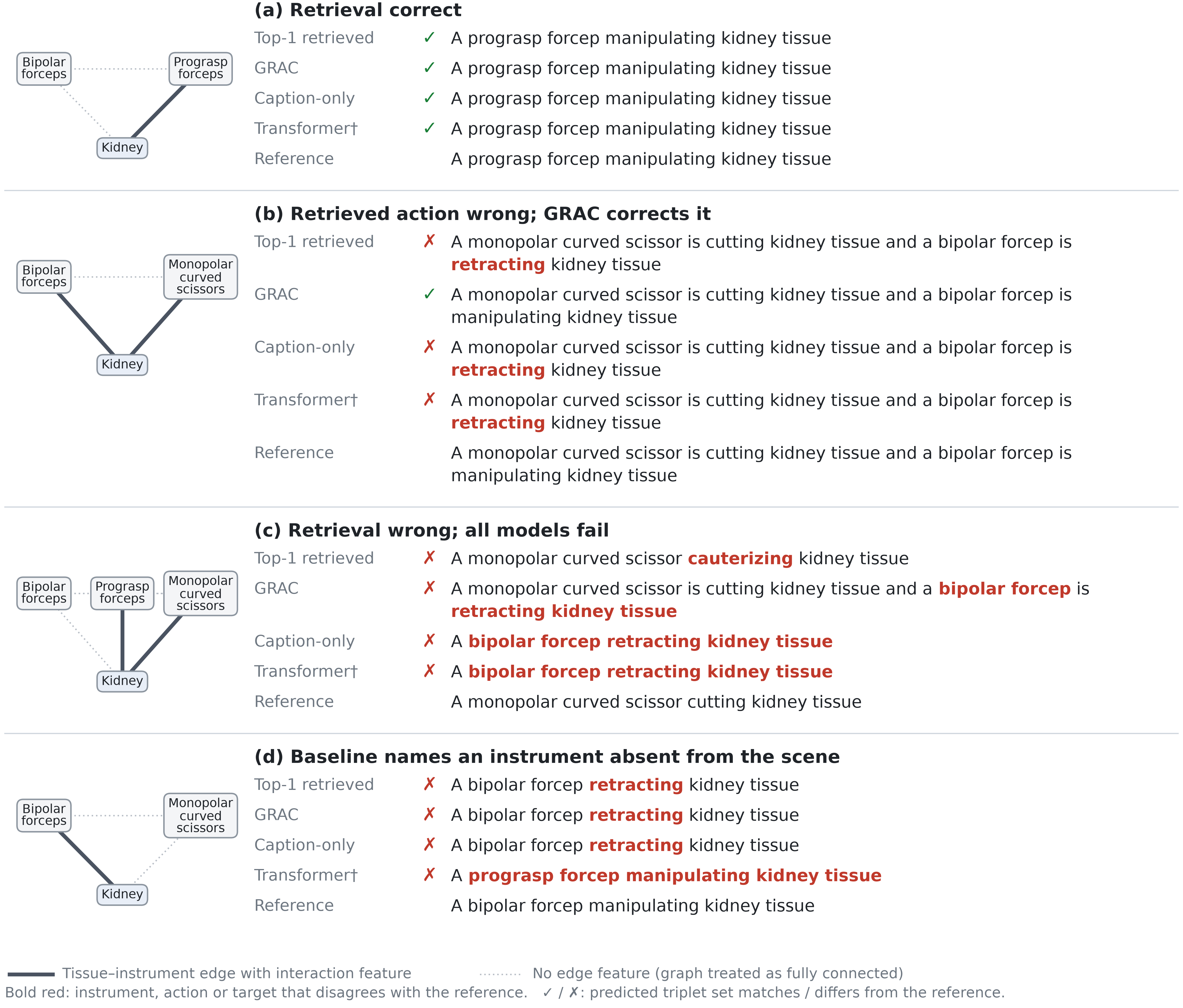}
\caption{Qualitative examples from the EndoVis-18 benchmark (sequences 1, 5 and 16). One frame was selected per category, at the median position among the frames that met the category definition: (a) correct Top-1 retrieval, (b) Top-1 retrieval with a wrong action that GRAC corrects, (c) wrong Top-1 retrieval with all models failing, and (d) a baseline naming an instrument absent from the scene. Bold red marks an instrument, action or target that disagrees with the reference; when the instrument is wrong, the whole triplet is marked. \cmark/\xmark\ indicate whether the predicted triplet set matches the reference.}\label{fig4}
\end{figure}

\textbf{Comparison with caption copying.} To assess whether caption generation provides benefits beyond retrieving previously observed descriptions, we evaluated a Top-1 Copy baseline that directly outputs the caption of GRAC's highest-ranked training example (Table~\ref{tab4}). Top-1 Copy achieved a CIDEr-D of 5.332 $\pm$ 0.150 and an exact triplet-set accuracy of 0.461 $\pm$ 0.012, compared with 5.883 $\pm$ 0.085 and 0.526 $\pm$ 0.008 for GRAC. The improvements were statistically significant (CIDEr-D: +0.552, 95\% CI [0.353, 0.760]; exact-set accuracy: +0.065, 95\% CI [0.044, 0.088]; Holm-adjusted p $<$ 0.001). The caption-only retrieval variant achieved a CIDEr-D of 5.805 $\pm$ 0.142, accounting for most of the improvement over copying. These findings indicate that retrieval of recurring surgical activities explains a substantial portion of the performance, while generation conditioned on the query graph provides additional benefits. Representative qualitative examples are shown in Fig.~\ref{fig4}.

\begin{table}[!htbp]
\caption{GRAC ablations on EndoVis-18 (392 frames). (a) Component configuration of each variant; a check mark denotes an enabled component and a dash an excluded component; the default input is frame $t$. (b) Language-generation results; values are mean $\pm$ SD over three seeds, and higher is better.}\label{tab3}
\centering
\footnotesize
\setlength{\tabcolsep}{2.6pt}
\renewcommand{\arraystretch}{1.15}
\begin{tabular*}{\textwidth}{@{\extracolsep{\fill}}lccccc@{}}
\toprule
\multicolumn{6}{@{}l}{\textit{(a) Component configuration}} \\
\midrule
\textbf{Variant} & \begin{tabular}[b]{@{}c@{}}\textbf{Edge}\\\textbf{features}\end{tabular} & \textbf{Retrieval} & \begin{tabular}[b]{@{}c@{}}\textbf{Graph}\\\textbf{comparison}\end{tabular} & \begin{tabular}[b]{@{}c@{}}\textbf{Compat.}\\\textbf{loss}\end{tabular} & \begin{tabular}[b]{@{}c@{}}\textbf{Query}\\\textbf{frames}\end{tabular} \\
\midrule
No retrieval or edges & -- & -- & -- & -- & $t$ \\
Graph only (with edges) & \checkmark & -- & -- & -- & $t$ \\
Caption retrieval & \checkmark & \checkmark & -- & -- & $t$ \\
Comparison, no compat.\ loss & \checkmark & \checkmark & \checkmark & -- & $t$ \\
Temporal context & \checkmark & \checkmark & \checkmark & \checkmark & $t-1,t,t+1$ \\
\textbf{GRAC (ours)} & \checkmark & \checkmark & \checkmark & \checkmark & $t$ \\
\bottomrule
\end{tabular*}
\par\vspace{8pt}
\begin{tabular*}{\textwidth}{@{\extracolsep{\fill}}lcccc@{}}
\toprule
\multicolumn{5}{@{}l}{\textit{(b) Language-generation results}} \\
\midrule
\textbf{Variant} & \textbf{BLEU-4} & \textbf{METEOR} & \textbf{ROUGE-L} & \textbf{CIDEr-D} \\
\midrule
No retrieval or edges & \grpm{0.651}{0.014} & \grpm{0.493}{0.018} & \grpm{0.778}{0.025} & \grpm{4.870}{0.450} \\
Graph only (with edges) & \grpm{0.699}{0.019} & \grpm{0.527}{0.006} & \grpm{0.811}{0.014} & \grpm{5.632}{0.090} \\
Caption retrieval & \grpm{0.710}{0.013} & \grpm{0.532}{0.016} & \grpm{0.814}{0.022} & \grpm{5.805}{0.142} \\
Comparison, no compat.\ loss & \grpm{0.709}{0.014} & \grpm{0.531}{0.013} & \grpm{0.818}{0.016} & \grpm{5.744}{0.198} \\
Temporal context & \grpmb{0.721}{0.009} & \grpm{0.532}{0.010} & \grpm{0.821}{0.013} & \grpm{5.749}{0.151} \\
\textbf{GRAC (ours)} & \grpm{0.715}{0.006} & \grpmb{0.541}{0.007} & \grpmb{0.828}{0.008} & \grpmb{5.883}{0.085} \\
\bottomrule
\end{tabular*}
\end{table}

\begin{table}[!htb]
\caption{Retrieval-only versus generated captions on EndoVis-18 (392 frames). Values are mean $\pm$ SD across three seeds; instrument hallucination is lower-is-better.}\label{tab4}
\centering
\footnotesize
\setlength{\tabcolsep}{2.6pt}
\renewcommand{\arraystretch}{1.15}
\begin{tabular*}{\textwidth}{@{\extracolsep{\fill}}lccccc@{}}
\toprule
\textbf{Method} & \textbf{CIDEr-D} & \textbf{BLEU-4} & \begin{tabular}[b]{@{}c@{}}\textbf{Exact}\\\textbf{set}\end{tabular} & \begin{tabular}[b]{@{}c@{}}\textbf{Triplet}\\\textbf{F1}\end{tabular} & \begin{tabular}[b]{@{}c@{}}\textbf{Instr.}\\\textbf{halluc. $\downarrow$}\end{tabular} \\
\midrule
Top-1 Copy & \grpm{5.332}{0.150} & \grpm{0.696}{0.013} & \grpm{0.461}{0.012} & \grpm{0.576}{0.013} & \grpm{0.060}{0.011} \\
Caption Retrieval & \grpm{5.805}{0.142} & \grpm{0.710}{0.013} & \grpm{0.520}{0.015} & \grpm{0.630}{0.013} & \grpm{0.049}{0.010} \\
\textbf{GRAC (ours)} & \grpmb{5.883}{0.085} & \grpmb{0.715}{0.006} & \grpmb{0.526}{0.008} & \grpmb{0.636}{0.012} & \grpmb{0.037}{0.015} \\
\bottomrule
\end{tabular*}
\end{table}

\subsection{Evaluation of a Second Dataset (TORS)}\label{subsec5.3}

To examine whether GRAC's performance advantage extends beyond EndoVis-18, we conducted an additional evaluation on the TORS dataset. All models were trained from scratch using the TORS training split, with checkpoint selection performed on the validation split, followed by retraining on the combined training and validation data. Each experiment was conducted with three random seeds, without TORS-specific hyperparameter tuning. Given the small validation set of 30 frames, minimum validation caption loss was used for checkpoint selection across all models. All methods received the released TORS scene-graph features, and performance was evaluated on 36 frames from three test sequences.

\begin{table}[!htb]
\caption{TORS test-set performance (36 frames). Values are mean $\pm$ SD over three seeds. (a) Language generation, including the CIDEr-D comparison with GRAC (GRAC minus baseline; paired frame-level bootstrap, 10,000 resamples; Holm-adjusted p-values). (b) Triplet-based factual agreement ($\downarrow$: lower is better).}\label{tab5}
\centering
\footnotesize
\setlength{\tabcolsep}{2.6pt}
\renewcommand{\arraystretch}{1.15}
\begin{tabular*}{\textwidth}{@{\extracolsep{\fill}}lccccc@{}}
\toprule
\multicolumn{6}{@{}l}{\textit{(a) Language generation}} \\
\midrule
\textbf{Model} & \textbf{BLEU-4} & \textbf{CIDEr-D} & \textbf{$\Delta$CIDEr-D} & \textbf{95\% CI} & \begin{tabular}[b]{@{}c@{}}\textbf{Adjusted}\\\textbf{p-value}\end{tabular} \\
\midrule
Transformer$^\dagger$ & \grpm{0.396}{0.011} & \grpm{3.80}{1.35} & $+1.26$ & $[0.31,\,2.25]$ & $0.010$ \\
M$^{2}$ Transformer$^\dagger$ & \grpm{0.363}{0.061} & \grpm{3.67}{1.69} & $+1.39$ & $[0.73,\,2.08]$ & $<0.001$ \\
AoA$^\dagger$ & \grpm{0.359}{0.045} & \grpm{3.61}{1.19} & $+1.46$ & $[0.86,\,2.11]$ & $<0.001$ \\
GET$^\dagger$ & \grpm{0.307}{0.019} & \grpm{1.97}{0.24} & $+3.09$ & $[1.51,\,4.70]$ & $<0.001$ \\
SGT & \grpm{0.387}{0.038} & \grpm{3.73}{1.46} & $+1.34$ & $[0.59,\,2.11]$ & $<0.001$ \\
\textbf{GRAC (ours)} & \grpmb{0.444}{0.031} & \grpmb{5.06}{0.26} & -- & -- & -- \\
\bottomrule
\end{tabular*}
\par\vspace{8pt}
\begin{tabular}{@{}lccc@{}}
\toprule
\multicolumn{4}{@{}l}{\textit{(b) Triplet-based factual agreement}} \\
\midrule
\textbf{Model} & \begin{tabular}[b]{@{}c@{}}\textbf{Exact}\\\textbf{set}\end{tabular} & \begin{tabular}[b]{@{}c@{}}\textbf{Triplet}\\\textbf{F1}\end{tabular} & \begin{tabular}[b]{@{}c@{}}\textbf{Instr.}\\\textbf{halluc. $\downarrow$}\end{tabular} \\
\midrule
Transformer$^\dagger$ & \grpm{0.352}{0.160} & \grpm{0.592}{0.038} & \grpm{0.102}{0.016} \\
M$^{2}$ Transformer$^\dagger$ & \grpm{0.333}{0.192} & \grpm{0.523}{0.157} & \grpm{0.222}{0.192} \\
AoA$^\dagger$ & \grpm{0.324}{0.143} & \grpm{0.508}{0.129} & \grpm{0.222}{0.147} \\
GET$^\dagger$ & \grpm{0.130}{0.016} & \grpm{0.331}{0.011} & \grpm{0.455}{0.039} \\
SGT & \grpm{0.343}{0.153} & \grpm{0.516}{0.143} & \grpm{0.245}{0.189} \\
\textbf{GRAC (ours)} & \grpmb{0.454}{0.016} & \grpmb{0.621}{0.013} & \grpmb{0.083}{0.028} \\
\bottomrule
\end{tabular}
\par\vspace{3pt}\raggedright\scriptsize
\end{table}

GRAC achieves the highest language-generation performance on TORS (Table~\ref{tab5}), obtaining a CIDEr-D score of 5.06 $\pm$ 0.26 and BLEU-4 of 0.444 $\pm$ 0.031. The strongest competing models are Transformer (CIDEr-D 3.80 $\pm$ 1.35) and SGT (3.73 $\pm$ 1.46). GRAC improves CIDEr-D over Transformer by 1.26 (95\% CI [0.31, 2.25], p = 0.010) and over SGT by 1.34 (95\% CI [0.59, 2.11], p $<$ 0.001). Significant improvements are also observed over the remaining baselines. GRAC's CIDEr-D also varied little across the three training seeds (SD 0.26, compared with 1.19--1.69 for Transformer, M\textsuperscript{2}, AoA, and SGT).

For factual agreement, GRAC achieves the highest exact triplet-set accuracy (0.454) and triplet F1-score (0.621), together with the lowest instrument hallucination rate (0.083). Compared with M\textsuperscript{2}, AoA, GET, and SGT, improvements across all three factual metrics are statistically significant (p $\leq$ 0.009). Relative to Transformer, the strongest competing model on these measures, the improvement in exact triplet-set accuracy is smaller (+0.102, p = 0.043), while differences in triplet F1 (+0.029) and instrument hallucination ($-$0.019) are not statistically significant.

The TORS evaluation provides complementary evidence that GRAC's relative advantage can be reproduced after independent training on a second surgical captioning dataset. Under the stated TORS protocol, GRAC has the highest mean language and triplet-based factual-agreement scores among the evaluated models. However, the evaluation comprises only 36 frames from three test sequences, and the reference captions are concentrated in recurring activities; the training set is likewise small. Accordingly, this result should be interpreted as within-dataset replication under a different surgical setting, not zero-shot transfer from EndoVis-18 or proof of robust generalization to independent clinical populations.

\section{Experiments}\label{sec4}

\subsection{Experimental Setup and Evaluation}\label{subsec4.1}

\textbf{Datasets.}
We evaluate GRAC on EndoVis-18~\cite{b2,b3,b8}, comprising 14 annotated robotic nephrectomy sequences with region-of-interest features, instrument--tissue interactions, and reference captions. Sequences 1, 5, and 16 (392 frames) form the held-out benchmark, while the remaining 11 sequences (1,124 frames) are divided into FIT (sequences 2, 3, 4, 6, 7, 10, 12, 14, and 15; 963 frames) and DEV (sequences 9 and 11; 161 frames). Training captions contain 56 distinct descriptions and 46 unique instrument--action--target triplet sets, with the five most frequent captions covering 55.6\% of frames. The training vocabulary contains 45 words. We additionally evaluate on TORS~\cite{b3,b4}, comprising 181 frames from 20 transoral robotic surgery sequences, using its predefined split of 115 training frames (14 sequences), 30 validation frames (sequences 1, 8, and 16), and 36 test frames (sequences 3, 10, and 17). The TORS training vocabulary contains 24 words.

\textbf{Training protocol.}
Hyperparameters and training duration are selected using the EndoVis-18 FIT/DEV split. GRAC and its ablations use minimum DEV caption cross-entropy for early stopping (patience 10; maximum 80 epochs), while baselines follow their respective training procedures with selection based on DEV CIDEr. All models are subsequently retrained from scratch on the 11 training sequences for their selected durations and evaluated using three independent seeds (1, 42, and 7). Benchmark data are excluded from all development decisions. On TORS, models are trained from scratch using the same architectures and hyperparameters without dataset-specific tuning.

\textbf{Evaluation metrics.}
Caption quality is assessed using BLEU-1--4, METEOR, ROUGE-L, and CIDEr-D through the COCO evaluation toolkit~\cite{b24,b25} (Java METEOR 1.5), following lowercasing and punctuation removal. CIDEr-D is reported on its native scale. For factual evaluation, a rule-based parser extracts instrument--action--target triplets~\cite{b26}, treating ``kidney'' and ``kidney tissue'' as equivalent targets across all methods. We report exact triplet-set accuracy, micro-averaged triplet precision, recall, and F1, and instrument hallucination, defined as the proportion of predicted instrument mentions absent from frame annotations. These metrics assess factual agreement but depend on parsing and annotation quality and do not directly quantify clinical error severity.

\textbf{Statistical analysis.}
Results are reported as mean $\pm$ standard deviation across three seeds. Statistical comparisons use 10,000 paired frame-level bootstrap resamples (392 EndoVis-18 and 36 TORS frames), with identical sampled indices across methods and seeds. Metrics are recomputed and averaged across seeds for each resample. We report 95\% percentile confidence intervals and two-sided p-values with Holm correction. Given within-sequence correlation, inference is limited to frame-level rather than procedure-level variability.

\begin{table}[htb]
\caption{GRAC architecture, optimization, and inference configuration.}
\label{tab6}
\centering
\footnotesize
\setlength{\tabcolsep}{2.6pt}
\renewcommand{\arraystretch}{1.05}
\begin{tabular*}{\textwidth}{@{\extracolsep{\fill}}p{0.30\linewidth}p{0.64\linewidth}@{}}
\toprule
\textbf{Component} & \textbf{Configuration} \\
\midrule
\multicolumn{2}{@{}l}{\textit{Architecture and retrieval}} \\
Graph encoder & 2 GATv2 layers; hidden dimension 512; 4 heads; ELU; dropout 0.057 \\
Query-memory encoder & 2 pre-norm Transformer layers; $d=384$; 4 heads; FFN 1,536; dropout 0.10 \\
Retrieved-caption encoder & 2 pre-norm Transformer layers; $d=384$; 4 heads; FFN 1,536; dropout 0.10 \\
Caption decoder & 2 Transformer layers; $d=384$; 4 heads; FFN 1,536; dropout 0.30 \\
Retrieval & 128-dimensional embeddings; $K=3$ distinct triplet sets; bank refreshed every 10 steps and after each epoch \\
\addlinespace[3pt]
\multicolumn{2}{@{}l}{\textit{Optimization}} \\
Retrieved-caption dropout & 0.3 per caption; 0.1 for all captions \\
InfoNCE temperature & $\tau=0.1$ \\
Label smoothing & 0.0197 \\
Optimizer & AdamW; weight decay $10^{-4}$ \\
Learning rates & $1.08\times10^{-4}$ (decoder backbone); $1.41\times10^{-4}$ (other modules) \\
Learning-rate schedule & Cosine annealing; $T_{\max}=80$ epochs; minimum LR $10^{-6}$ \\
Batch size & 32 captioning queries + 64 contrastive scenes \\
Gradient clipping & 1.0 \\
Model selection & Minimum validation caption cross-entropy; maximum 80 epochs; patience 10 \\
\addlinespace[3pt]
\multicolumn{2}{@{}l}{\textit{Inference and resources}} \\
Decoding & Beam width 2; length normalization 0.8; maximum 20 tokens \\
Trainable parameters & 15,932,143 (15.9 M) \\
Hardware/software & NVIDIA Tesla V100 (32 GB); PyTorch 2.5.1 \\
\bottomrule
\end{tabular*}
\end{table}

\section{Discussion}\label{sec6}

The results demonstrate that GRAC benefits from combining relational graph representations with activity-guided retrieval for surgical description generation. The ablation and error analyses further reveal distinct contributions from graph encoding, retrieved descriptions, and supervised graph comparison, particularly in improving factual agreement.

\subsection{Graph Encoding and Activity-Guided Retrieval}\label{subsec6.1}

The ablation results show that instrument--tissue interactions and retrieved captions make complementary contributions to caption generation. Incorporating edge features increases CIDEr-D from 4.870 to 5.632, indicating the importance of relational information beyond individual node representations. Adding caption retrieval further improves performance to 5.805, suggesting that previously observed surgical descriptions provide useful contextual information when similar activities recur across sequences.

Importantly, these gains cannot be explained by direct caption copying alone. Reproducing the highest-ranked retrieved caption achieves CIDEr-D of 5.332 and exact triplet-set accuracy of 0.461, compared with 5.883 and 0.526 for GRAC. The improvement indicates that combining retrieved descriptions with query-specific graph information provides benefits beyond simply reusing the closest caption. Nevertheless, the relatively strong copying baseline highlights the influence of recurring activities and the restricted caption vocabulary.

Additional experiments indicate that providing more context does not necessarily improve caption generation. Increasing the number of retrieved examples beyond three reduced performance, while incorporating neighboring frames offered no improvement over the single-frame configuration. These findings suggest that the relevance of contextual information may be more important than its quantity in the evaluated setting.

\subsection{Graph Comparison and Factual Reliability}\label{subsec6.2}

The graph-comparison results reveal a distinction between improvements in language quality and factual grounding. Adding comparison features without compatibility supervision reduces CIDEr-D to 5.744, below caption-only retrieval (5.805). The complete model achieves 5.883, although its incremental CIDEr-D improvement over caption-only retrieval is not statistically significant ($p=0.054$). In contrast, instrument hallucination decreases significantly from 0.049 to 0.037 ($p=0.003$). This suggests that compatibility supervision helps the model use comparison features, with its clearest observed benefit being improved instrument-level factual agreement rather than a substantial increase in aggregate caption quality.

The error analysis provides further insight into this effect. GRAC corrected 26 instrument errors compared with eight regressions, indicating that graph-conditioned caption representations can help prevent incorrect instrument mentions. However, correcting action errors remains challenging. When the highest-ranked retrieved caption contains the correct instruments but an incorrect action, GRAC reproduces that action in 54--71\% of the examined frames. Thus, successful instrument correspondence does not necessarily ensure correct interpretation of the associated surgical activity.

These findings highlight that improved retrieval and instrument-level grounding do not necessarily ensure correct action descriptions, emphasizing the need for more reliable alignment between retrieved surgical interactions and the query scene.

\subsection{Evaluation and Generalizability}\label{subsec6.3}

GRAC outperforms the evaluated baselines under the common experimental protocol, including the reproduced SGT model (CIDEr-D: 5.883 versus 4.234). However, comparisons with published SGT and SGT++ results require consideration of differences in input features and evaluation procedures. The published SGT protocols involves benchmark-based model selection, whereas GRAC's primary training procedure used development-based checkpoint selection.  The lack of a complete public SGT++ implementation and matching processed features further limits direct comparison with its published results.

An additional evaluation on TORS supports the applicability of GRAC's approach to another surgical setting, achieving CIDEr-D of 5.06 and exact triplet-set accuracy of 0.454. However, this experiment involved independent training rather than cross-dataset transfer and included only 36 test frames from three sequences.

\subsection{Practical Implications and Future Directions}\label{subsec6.4}

Generating concise descriptions of instrument--tissue interactions could support surgical video indexing, activity-based retrieval, and procedural review. By combining retrieved linguistic context with scene-specific relational information, GRAC provides a framework for generating descriptions grounded in structured surgical observations. However, its current output consists of frame-level descriptions rather than complete operative reports.

Despite improved captioning performance, GRAC's evaluation remains limited by repetitive surgical activities, modest factual accuracy, and small procedure-specific datasets. Further validation on diverse surgical procedures and previously unseen activities is needed to establish its generalizability and clinical reliability. Future work should evaluate the framework using automatically predicted scene graphs and more diverse surgical procedures, with expert assessment of factual accuracy and clinical relevance.

\section{Conclusion}\label{sec7}

This study presents GRAC, a graph-based retrieval-augmented framework that improves surgical scene captioning performance on EndoVis-18 and TORS. The results demonstrate the complementary benefits of interaction-aware graph encoding and caption retrieval, while compatibility-supervised graph comparison further reduces instrument hallucination. These findings highlight the potential of structured retrieval for generating more accurate and factually consistent surgical scene descriptions.

\section*{Declarations}

\noindent\textbf{Code and data availability.} The code used to generate the reported results will be made publicly available at \url{https://github.com/MIILab-MTU/SurgicalVideoCaptioningByGRAC} upon publication. EndoVis-18 and TORS are publicly available from their original sources.

\noindent\textbf{Funding.} This research was supported in part by grants from the National Institutes of Health (1R15HL172198 and 1R15HL173852) and the American Heart Association (25AIREA1377168).

\noindent\textbf{Conflict of interest.} Dr. Weihua Zhou reports consulting fees from SynterMed Inc., outside the submitted work. All other authors declare no competing interests.

\noindent\textbf{Ethics approval.} This study used only publicly available, de-identified datasets; no new data were collected, and ethics approval was not required.

\end{document}